\pdfoutput=1

\documentclass[11pt]{article}
\usepackage{acl}
\usepackage{times}
\usepackage{latexsym}
\usepackage[T1]{fontenc}
\usepackage[utf8]{inputenc}
\usepackage{microtype}
\usepackage{inconsolata}
\usepackage{graphicx}
\usepackage{amsmath}
\usepackage{booktabs}
\usepackage{CJKutf8}

\title{Measuring Information Distortion  in Hierarchical Ultra long Novel Reconstruction:\\The Optimal Expansion Ratio}

\author{HANWEN SHEN\\Stevens Institute of Technology \\ \texttt{hshen13@stevens.edu} \And TING YING }

\begin{document}
\begin{CJK}{UTF8}{gbsn}
\maketitle


\begin{abstract}

A two stage novel generation framework (outline -> section outline -> manuscript) is widely used in long novel generation,(e.g., \textsc{DOME}, \textsc{Plan\&Write}, \textsc{Long Writer}), but study of such framework in ultra long novel(>1M words) reconstruction is little.

Building on recent text compression methods (\textsc{LLMZip}, \textsc{LLM2Vec}), we conduct an information-theoretic analysis to quantify semantic distortion under different compression–expansion ratios. We  examine how outline length affects information preservation.

Experiments on ultra long novels show that the optimal compression-expansion ratio significantly reduce semantic distortion compare to other non-optimal compression-expansion ratio.

\end{abstract}

\section{Introduction}
We observe a fundamental phenomenon in large-scale novel generation \citep{PLMTextGenSurvey2025,SuContrastiveGen}: when a million-word novel is summarized into a 1,000-word outline by an LLM (e.g., ChatGPT) and then expanded, the result shows substantial distortion, detail loss, and semantic divergence. In contrast, compressing a 100k-word novel into a 10k-word outline and expanding it preserves meaning more faithfully.

Studies such as LongWriter agree that generating 100k+ word novels typically follows an outline-to-novel workflow. The quality depends on two prompt-side variables: (i) outline length, and (ii) detail density (e.g., characters, plots, scenes). A trade-off emerges: fewer tokens and sparse detail reduce quality, while rich, lengthy prompts shift the burden to the human author.

Our ultimate goal is to generate detail-rich ultra-long novels from minimal outlines. However, since "detail" itself is hard to define and evaluate directly, we instead study the detail loss in reconstruction of ultra-long novels under varying compression ratios using large language models, as a proxy for generation quality.

\subsection{Motivation}
Ultra-long novels (1M+ words) are highly popular on \textit{WuxiaWorld}, \textit{Fanqie}, and \textit{Qidian}, making automated generation a key goal. Since ChatGPT, models like LLaMA \citep{Grattafiori2024Llama3}, DeepSeek \citep{DeepSeek2025TechnicalReport}, Qwen \citep{Bai2023Qwen}, Gemini \citep{TeamG2024Gemini}, and GPT-4o \citep{OpenAI2024GPT4o} have advanced long-context processing.

However, despite 1M-token input capacity, output limits (e.g., 16k) make faithful reconstruction difficult \citep{Mikhaylovskiy2023LongTextGen}.
Although many prior studies have addressed novels up to 100k words, empirical evidence is lacking to show that methods effective at this scale naturally extend to generating novels of 1 million words. Challenges at the million-word scale—maintaining coherence, thematic consistency, and character development—are qualitatively and quantitatively different. Inspired by the encoder–decoder paradigm, we adopt a reconstruction-based framework as a surrogate objective to study and improve ultra-long text generation.

\begin{table}[htp!]
\centering
\tiny
\begin{tabular}{lccc}
\toprule
\textbf{Model} & \textbf{Context Size} & \textbf{Max TPM} & \textbf{Max Output}\\
\midrule
Gemini 2.0 Flash & 1,000,000 & 1,000,000 & 8,192 \\ 
Claude 3.7 Sonnet & 200,000 & 200,000 & 128,000 \\ 
GPT-4.1 & 1,000,000 & 400,000 & 16,000 \\ 
Chatgpt-4o & 128,000  & 800,000	 & 16,384 \\ 
LLAMA 4 Scout & 10,000,000 & 1,000,000 & 8,000 \\
OPENAI O3 & 200,000 & 200,000 & 100,000 \\
\bottomrule
\end{tabular}
\caption{Comparison of large language models by context size, maximum tokens processed in a minute, and maximum output length in tokens. On average, one token corresponds to roughly 3 words in English, 2 in Chinese. We quote the tier 3 TPM limit for GPT-4.1, OPENAI O3, and tier 3 TPM limit for Claude 3.7 Sonnet. Data are collected from their websites.}
\label{tab:model-comparison}
\end{table}

\begin{figure*}[!t]
    \centering
    \includegraphics[width=1\textwidth]{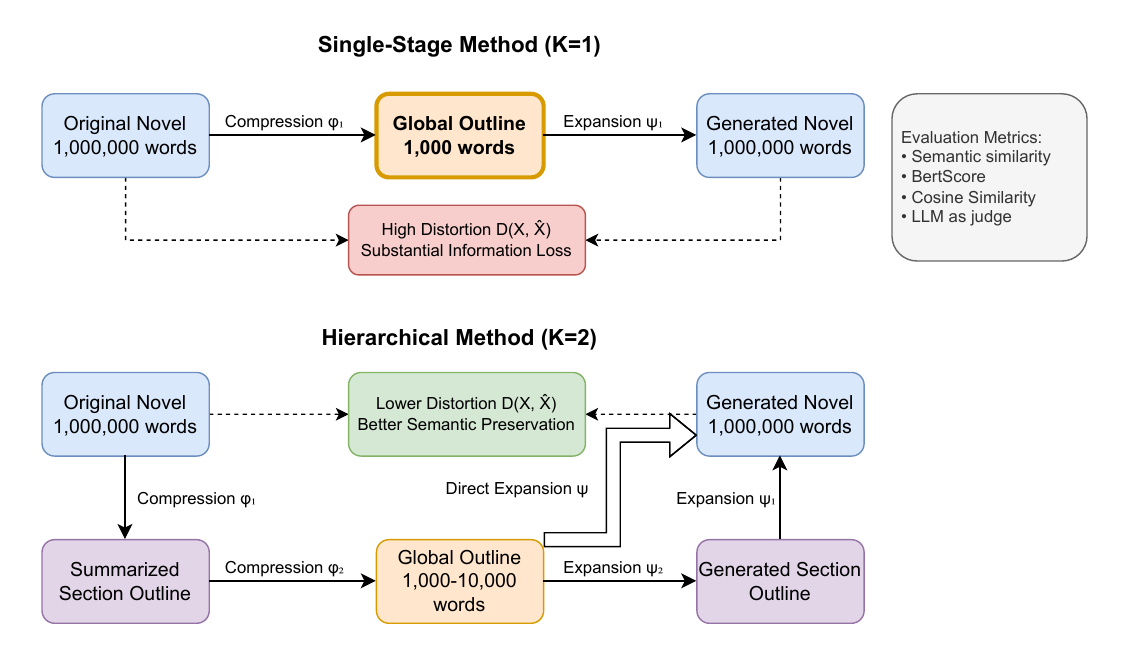}
    \caption{Pipeline for ultra-long novel generation using hierarchical outline approach. Our method maintains narrative fidelity by optimizing compression-expansion ratios across multiple stages.}
    \label{fig:pipeline}
\end{figure*}




\subsection{Contribution} 
\begin{itemize} 
\item We establish a quantitative relationship between outline compression-expansion ratios and distortion (detail loss) through experiments. \(R=0.01\) is the optimal compression-expansion ratio under our configuration and experiment.
\item We propose and empirically verify that mixed two-stage hierarchical outline decreases distortion in generated novels compared to traditional one-stage and two-stage approaches. 
\end{itemize}

\subsection{Related Work}

\paragraph{Long-text Generation.}
Long-text generation introduces challenges in coherence, planning, and discourse modeling \citep{Kumar2024LongLaMP,Que2024HelloBench,Wu2025LongEval}. Techniques include retrieval-augmented generation \citep{Lewis2020RAG}, self-refinement \citep{Madaan2023SelfRefine}, and iterative lengthening \citep{Quan2024SelfLengthen}. Novel generation demands plot and character consistency over long spans \citep{Xie2023NextChapter,Yao2019PlanWrite,Guan2021LongText}. Hierarchical planning \citep{Wang2023Pacing,Yang2023DOC,Wang2024DynamicOutlining}, memory-based agents \citep{Bai2024LongWriter}, and reinforcement-based control \citep{SciAdv2024Creativity,ChhunMetaEval2025} have been explored, but ultra-long contexts remain difficult.

\paragraph{LLM-based compression.}
Some work treats LLMs as adaptive entropy coders \citep{Shin2023TextCompression}, e.g., \textsc{LLMZip} \citep{Valmeekam2023LLMZip} and \textsc{DeepZip} \citep{Goyal2018DeepZip}, achieving near-Shannon compression via token probability coding. We instead explore lossy but semantically faithful compression: can an LLM expand a sparse outline into a coherent million-word novel? We analyze the acceptable range of outline entropy $H_{\text{outline}}$ that balances compression with reconstructive fidelity.

\paragraph{LLM as judge.}
We assess both (i) distortion from the original novel and (ii) the quality of generated output. Distortion metrics include traditional algorithms (Cosine, Hamming, Jaccard, Levenshtein, TF-IDF) \citep{Gahman2023IJAI,SurveyTextSimilarityRG,Chatterjee2020NewsSimilarity,WangDong2020SimilaritySurvey,Gahman2023SimilarityArXiv} and pre-trained models (BERTScore, BLEU, ROUGE, METEOR) \citep{Guan2022LOT,Xiao2024CPack,Yu2024PatentSimilarity,Yuan2023CTSARF}. Novel quality is evaluated via aesthetic score, coherence, creativity, consistency, and fluency \citep{Ichmoukhamedov2024XAIStories,Venkatraman2024GPTWho}, with recent studies proposing LLMs themselves as evaluators \citep{Gu2025LLMJudge}.


\section{Methodology}

\subsection{Notation and Two-Stage Pipeline.}
Let \(X_0\in\mathcal{V}^{L_0}\) denote the original novel with \(L_0\) words.
A sequence of \(K\) compression mappings
\[
  \{\phi_i:\mathcal{V}^{L_{i-1}}\!\rightarrow\!\mathcal{V}^{L_i}\}_{i=1}^{K}
\]
produces progressively shorter outlines
\[
  X_i=\phi_i(X_{i-1}),\qquad
  \alpha_i=\frac{L_i}{L_0},\qquad i=1,\dots,K,
\]
where \(X_i\in\mathcal{V}^{L_i}\) and \(\alpha_i\) is the cumulative compression ratio after level \(i\).

Restoration applies the corresponding expansion mappings
\[
  \{\psi_i:\mathcal{V}^{L_i}\!\rightarrow\!\mathcal{V}^{L_{i-1}}\}_{i=K}^{1}:
  \quad
  \hat{X}_{i-1}=\psi_i(\hat{X}_i),
\]
with initials
\[
  \hat{X}_K=X_K,\qquad
  \hat{X}=\hat{X}_0.
\]
The overall objective, following rate–distortion theory\citep{Blau2019Rethinking}, is
\[
  \min_{\{\phi_i,\psi_i\}_{i=1}^{K}}
  \;R=\frac{H(X_K)}{L_0}
  \quad\text{s.t.}\quad
  D(\hat{X},X_0)\le\epsilon,
\]
where \(H(\cdot)\) is defined as the number of words of text\citep{Bentz2017} and \(D(\cdot,\cdot)\) is a task-specific distortion bounded by \(\epsilon\) in our study. 

\medskip\noindent
\paragraph{Outline–Expansion Pipeline.}
We implement three pipelines, as shown in Figure~\ref{fig:pipeline}:
\textbf{(i)Single-Stage:} 
generated from the full novel, then expanded directly to explore distortion under extreme compression.
\textbf{(ii)Hierarchical Two-Stage:} The compression first reduces the original novel \(X_0\) to a high-level outline \(X_1\) with rate \(\alpha_1\), and then further compresses it to a global summary \(X_2\) with rate \(\alpha_2\). The expansion reconstructs \(\hat{X}_1\) from \(X_2\), and finally recovers \(\hat{X}_0\) from \(\hat{X}_1\).

\textbf{(iii)Mixed Two-Stage:} The compression is the same as above. However, the expansion skips the intermediate outline and directly reconstructs \(\hat{X}_0\) from \(X_2\), under varying compression ratios \(R\).


We can derive the compression ratio to be \( R = \alpha_1 \times \alpha_2 \).

Using traditional methods like LZ77, lossless compression achieves \(R \approx 0.5\) \citep{amir_et_al:LIPIcs.CPM.2024.3}, which serves as a lower bound.

\textbf{We define distortion as a composite of Cosine distance, BERTScore, and Euclidean distance of detail counts, detailed in our evaluation metrics.}

\begin{table*}[ht]
\centering
\tiny
\setlength{\tabcolsep}{5pt}
\begin{tabular}{lccccccccc}
\toprule
\textbf{Setting} & $\mathbf{R = \alpha_1 \times \alpha_2}$ & \textbf{Cosine} & \textbf{BERT F1} & \textbf{SemSim} & \textbf{CharSim} & \textbf{StyleSim} & \textbf{CharDiff} & \textbf{SceneDiff} & \textbf{PropDiff} \\
\midrule
B & 1.000 & 0.922 $\pm$ 0.054 & 0.602 $\pm$ 0.085 & 0.912 $\pm$ 0.081 & 0.914 $\pm$ 0.104 & 0.919 $\pm$ 0.096 & 0.50 $\pm$ 1.18 & 0.36 $\pm$ 0.99 & 0.44 $\pm$ 1.00 \\
C & 0.010 & 0.645 $\pm$ 0.069 & 0.169 $\pm$ 0.038 & 0.418 $\pm$ 0.179 & 0.410 $\pm$ 0.195 & 0.350 $\pm$ 0.160 & 7.36 $\pm$ 5.77 & 3.85 $\pm$ 3.75 & 4.65 $\pm$ 4.89 \\
D & 0.010 & 0.633 $\pm$ 0.077 & 0.160 $\pm$ 0.045 & 0.390 $\pm$ 0.209 & 0.420 $\pm$ 0.202 & 0.305 $\pm$ 0.191 & 7.77 $\pm$ 4.96 & 5.11 $\pm$ 3.38 & 6.24 $\pm$ 5.05 \\
K2-* & 0.010 & \textbf{0.677 $\pm$ 0.073} & \textbf{0.199 $\pm$ 0.046} & \textbf{0.613 $\pm$ 0.216} & \textbf{0.566 $\pm$ 0.201} & \textbf{0.611 $\pm$ 0.133} & \textbf{6.70 $\pm$ 5.89} & \textbf{4.13 $\pm$ 4.07} & \textbf{4.66 $\pm$ 5.27} \\
K2-0 & 0.001 & 0.637 $\pm$ 0.068 & 0.169 $\pm$ 0.040 & 0.439 $\pm$ 0.201 & 0.289 $\pm$ 0.164 & 0.493 $\pm$ 0.139 & 10.19 $\pm$ 6.53 & 5.10 $\pm$ 3.79 & 4.88 $\pm$ 5.08 \\
K2-1 & 0.005 & 0.661 $\pm$ 0.077 & 0.189 $\pm$ 0.047 & 0.591 $\pm$ 0.234 & 0.474 $\pm$ 0.189 & 0.600 $\pm$ 0.147 & 8.62 $\pm$ 5.75 & 4.65 $\pm$ 4.06 & 4.88 $\pm$ 4.63 \\
K2-2 & 0.010 & 0.665 $\pm$ 0.074 & 0.188 $\pm$ 0.047 & 0.578 $\pm$ 0.232 & 0.488 $\pm$ 0.202 & 0.581 $\pm$ 0.145 & 8.70 $\pm$ 6.04 & 4.63 $\pm$ 3.74 & 5.16 $\pm$ 5.87 \\
K2-3 & 0.001 & 0.650 $\pm$ 0.070 & 0.168 $\pm$ 0.041 & 0.517 $\pm$ 0.210 & 0.354 $\pm$ 0.183 & 0.538 $\pm$ 0.137 & 9.88 $\pm$ 6.26 & 4.97 $\pm$ 3.84 & 5.02 $\pm$ 4.88 \\
K2-4 & 0.005 & 0.649 $\pm$ 0.074 & 0.179 $\pm$ 0.038 & 0.572 $\pm$ 0.217 & 0.452 $\pm$ 0.169 & 0.575 $\pm$ 0.134 & 9.30 $\pm$ 6.26 & 4.81 $\pm$ 3.97 & 5.20 $\pm$ 5.61 \\
K2-5 & 0.010 & 0.655 $\pm$ 0.074 & 0.181 $\pm$ 0.043 & 0.549 $\pm$ 0.231 & 0.458 $\pm$ 0.192 & 0.572 $\pm$ 0.147 & 9.15 $\pm$ 6.38 & 4.87 $\pm$ 3.92 & \textbf{4.66 $\pm$ 5.72} \\
K2-6 & 0.015 & 0.668 $\pm$ 0.073 & 0.193 $\pm$ 0.041 & 0.667 $\pm$ 0.184 & 0.535 $\pm$ 0.167 & 0.628 $\pm$ 0.119 & 8.76 $\pm$ 6.56 & 4.98 $\pm$ 3.94 & 5.03 $\pm$ 6.07 \\
K2-7 & 0.020 & 0.668 $\pm$ 0.07 & 0.192 $\pm$ 0.044 & 0.71 $\pm$ 0.166 & 0.583 $\pm$ 0.158 & 0.654 $\pm$ 0.110 & 9.53 $\pm$ 7.58 & 5.09 $\pm$ 4.30 & 5.42 $\pm$ 7.00 \\
K2-8 & 0.015 & 0.674 $\pm$ 0.075 & 0.198 $\pm$ 0.043 & 0.67 $\pm$ 0.181 & 0.531 $\pm$ 0.168 & 0.634 $\pm$ 0.116 & 8.47 $\pm$ 6.59 & 4.63 $\pm$ 3.81 & 4.49 $\pm$ 4.11 \\
K2-9 & 0.020 & 0.679 $\pm$ 0.073 & 0.197 $\pm$ 0.045 & 0.663 $\pm$ 0.193 & 0.543 $\pm$ 0.202 & 0.633 $\pm$ 0.125 & 9.41 $\pm$ 6.23 & 5.17 $\pm$ 4.15 & 5.31 $\pm$ 5.03 \\
\bottomrule
\end{tabular}
\caption{
Grouped statistics (mean $\pm$ std) for similarity and structural difference metrics, excluding the baseline (B), to identify best-performing configurations. 
The compression ratio is computed as \( R = \alpha_1 \times \alpha_2 \). 
Bolded values indicate the best average performance for each metric among the tested configurations under $R\leq 0.01$, but do not imply statistical significance. 
Formal significance testing results are presented in a separate table. Distortion is computed as one minus similarity.
}

\label{tab:r-only-summary}
\end{table*}

\subsection{Dataset Configuration}

We sample 40 public-domain Chinese novels (10 each from Fantasy, Urban, Romance, and Historical genres) from dataset \citep{webnovel_cn}. Each 1M-word novel is split into ~5,000-word chapters for reliable processing.

Compression/expansion prompts are detailed in Appendix~\ref{se:Appendixbaseline}–\ref{se:AppendixHierarchicalPrompts}. Translation experiments confirm language versions have minimal impact.

For evaluation (Appendix~\ref{sec:design}), we generate 200 section and one global outline per novel. Eight chapters per novel are reconstructed for comparison.

\subsection{Controlled Variables.}
Our configuration space is defined by five independent factors:  
(i) the \textit{novel genre} selected for evaluation;   
(ii) the \textit{compression ratio} applied at each outline level (global and section);  
(iii) the \textit{LLM model}, which is the backbone model.


\subsection{Evaluation Metrics}

To quantify reconstruction distortion, we define a three distance function \(D(X, \hat{X})\) as:

  

 \(D_{\text{trad}}(X, \hat{X})\) combines one minus the cosine similarity score and one minus BERTScore between embeddings of original and reconstructed novels; 
  
   \(D_{\text{llm}}(X, \hat{X})\) is the average GPT-4o judgment score across semantic, plot, character, background, and style similarity;
  
   \(D_{\text{struct}}(X, \hat{X})\) measures absolute differences in counts of unique characters, scenes, and items.





\section{Experiment}

We evaluate hierarchical compression–expansion strategies for ultra-long novel generation using \textbf{Gemini 2.0 Flash} (temperature = 0.3), focusing on:  
(1) How do different compression ratios affect semantic fidelity?  
(2) Does a two-stage outline hierarchy (\(K=2\)) outperform a single-stage (\(K=1\))?

Gemini is chosen for its large context window and throughput efficiency. Each novel involves near 3M input and 500K output tokens.

\subsection{Baseline Setup ($K = 1$)}

We test a single-stage pipeline using a \textbf{10,000-word outline} (\(R=\alpha_1 \approx 0.01\)) generated from the full novel, then expanded directly to explore distortion under extreme compression.

We compare four methods:  
\textbf{(A)} Human-written original,  
\textbf{(B)} Translate → Reconstruct,  
\textbf{(C)} Direct compression–expansion, and  
\textbf{(D)} LongWriter.

\subsection{Hierarchical Setup ($K = 2$)}

In the two-stage pipeline, the novel \(X_0\) is compressed into a section-level outline \(X_1\), then further into a global outline \(X_2\), enabling multi-resolution control.

Compression ratios \(\alpha_1 \in \{0.05, 0.10\}\), \(\alpha_2 \in \{0.01, 0.05, 0.10, 0.20, 0.30, 0.40\}\) determine abstraction levels (Table~\ref{tab:gemini-k2-config}).

\begin{table}[h]
\centering
\small
\begin{tabular}{lcccc}
\toprule
\textbf{ID} & $\alpha_1$ & $\alpha_2$ & $L_1$  & $L_2$  \\
\midrule
K2-0 & 0.05 & 0.01 & 50,000 & 1,000 \\
K2-1 & 0.05 & 0.10 & 50,000 & 5,000 \\
K2-3 & 0.05 & 0.20 & 50,000 & 10,000\\
K2-4 & 0.10 & 0.01 & 100,000 & 1,000 \\
K2-5 & 0.10 & 0.05 & 100,000 & 50,000 \\
K2-6 & 0.05 & 0.30 & 50,000 & 15,000 \\
K2-7 & 0.05 & 0.40 & 50,000 & 20,000 \\
K2-8 & 0.10 & 0.15 & 100,000 & 15,000 \\
K2-9 & 0.10 & 0.20 & 100,000 & 20,000 \\
K2-* & 0.05 & 0.20 & 50,000 & 10,000 \\
\bottomrule
\end{tabular}
\caption{Two-stage compression configurations under Gemini-2.0-Flash. The method K2-* is first apply two stage compressions, then apply direct expansion.}
\label{tab:gemini-k2-config}
\end{table}

Outlines is structured using JSON templates paragraph summaries.

\section{Analysis}

\subsection{Significance Testing}

Pairwise $t$-tests show that most compression configurations differ significantly in distortion ($p < 0.05$), especially in character similarity. Notably, \textbf{K2-*} outperforms all settings with $R \leq 0.01$ ($p < 0.001$), while \textbf{K2-3} underperforms compared to \textbf{K2-4} and \textbf{K2-5}. Some comparisons (e.g., K2-1 vs.\ K2-2, $p = 0.60$) show no significant difference, suggesting robustness in certain ranges. \textbf{B} confirms that translated text remains semantically close to the original.

Distortion decreases with higher compression ratio, but gains from $R = 0.001$ to $0.01$ are larger than those from $0.01$ to $0.02$.

\paragraph{Visualization.} Appendix figures show: (i) correlation between $R$ and similarity (Figure~\ref{fig:r-vs-mean-similarity}); (ii) grouped statistics (Table~\ref{tab:gemini-k2-config}); (iii) significance heatmap (Figure~\ref{fig:significance-heatmap}).

\paragraph{Sampling Justification.}

Although each novel has about 200 chapters, we sample 8 chapters per novel. A pilot study over 40 books shows a Pearson correlation \(r=0.95\) between full-book distortion and sampled distortion.

Fisher \(z\)-transform analysis confirms that \(r > 0.90\) even under 95\% confidence, validating the proxy method.

\section{Conclusion}

Using \textbf{Gemini 2.0 Flash}, we find that structured JSON outlines with \(\alpha_1 = 0.05\), \(\alpha_2 = 0.20\), and direct expansion (\textbf{K2-*}) yield the best fidelity to original novels in both semantic and structural metrics.

\section{Discussion}

We hypothesize that section-level outlines improve compression by guiding localized abstraction, but may be less necessary in expansion due to the global outline’s context coverage. The weak correlation between $R$ and similarity likely reflects the model's ability to exploit global context, reducing reliance on fine-grained section summaries.

\section*{Limitations}
This study has several limitations. First, the proposed hierarchical framework may face challenges when applied to novels with intricate or nonlinear structures, such as mysteries. Second, all experiments were conducted on Chinese texts, which may limit the generalizability of our findings across cultural contexts—a factor that extends beyond language alone. Third, the framework incurs substantial computational costs, potentially constraining its scalability. Selecting $H(\cdot)$ to be the number of words of text is a practical way compared to computing Shannon entropy of text. 
Finally, although care was taken to ensure fair evaluation, the use of LLM-based scoring may introduce systematic biases. In practical applications, human evaluation may serve as a more accurate and context-aware assessment method.

During compression and reconstruction, we acknowledging that all two stages   incur token costs. However, we exclude \( \sum_{k=1}^{2} H(X_k) = \alpha_1 \times \alpha_2 + \alpha_1\) from consideration, as our focus is on generating ultra-long novels from global outlines. We do not fix a specific value of 
$\epsilon$, as our study demonstrates that differences in performance across compression ratios are significant. However, there is insufficient evidence to justify any particular choice of 
$\epsilon$ as universally optimal.

\section*{Ethical Considerations}
We exclusively chose publicly available sources \citep{webnovel_cn} and Chinese version of ultra long novels from Project Gutenberg for evaluation. We follow the claim 'This dataset and any derivatives generated from it may be used for research purposes only. Commercial use and any other applications that may cause harm to society are strictly prohibited.' Any data contains offensive content has been filtered by Gemini 2.0 flash.
Given our result will be an optimal hyperparameter and no pretrained model or dataset will be provided to the public, the risk of ethical concerns is minimal.
However, we should also consider that the use of language models in long-form creative writing may impact authors’ livelihoods and raise concerns about bias and the propagation of misinformation.

\section*{Acknowledgments}
We gratefully acknowledge the use of AI-assisted tools solely for grammatical corrections during manuscript preparation. No other aspects of the research— including conceptualization, experimental design, data analysis, or interpretation of results—were generated or modified by AI. All substantive content and conclusions were developed independently by the authors.

\section*{Implementation Details}
Code has been published at 
\href{https://anonymous.4open.science/r/Measuring-Information-Entropy-in-Hierarchical-Ultra-long-Novel-Generation-38EA/}{anonymous space}.

\bibliography{main}
\clearpage
\appendix

\section{Appendix A: Pairwise significance test results between compression configurations}

\begin{figure}[h]
    \centering
    \includegraphics[width=\textwidth]{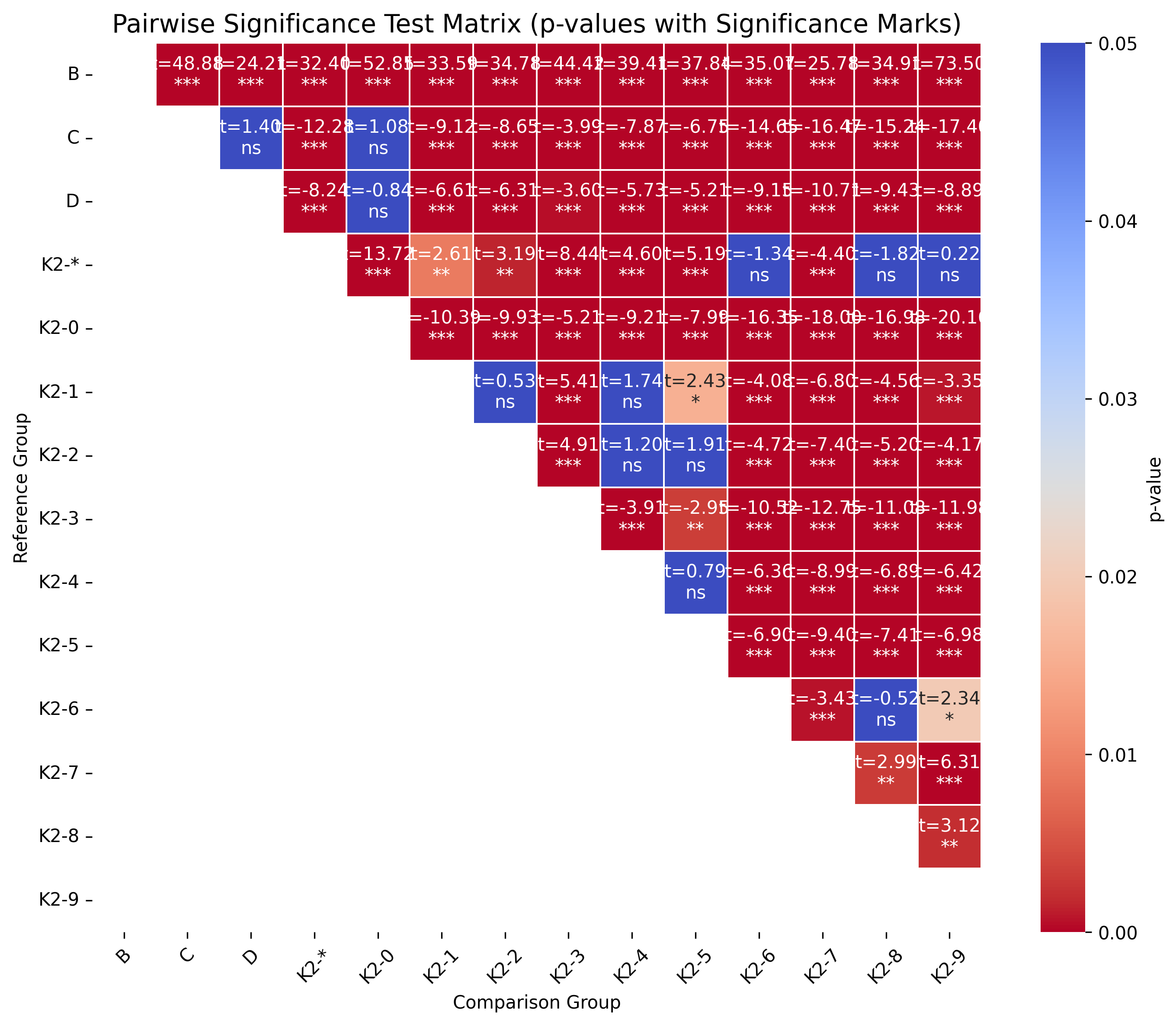}
    \caption{Pairwise significance test results between compression configurations. Each cell shows the $t$-value and significance level (* for $p<0.05$, ** for $p<0.01$, *** for $p<0.001$).}
    \label{fig:significance-heatmap}
\end{figure}

\clearpage
\section{Appendix B:Correlation Between R and Mean Similarity Excluding Group BD }
\begin{figure}[h]
    \centering
    \includegraphics[width=\textwidth]{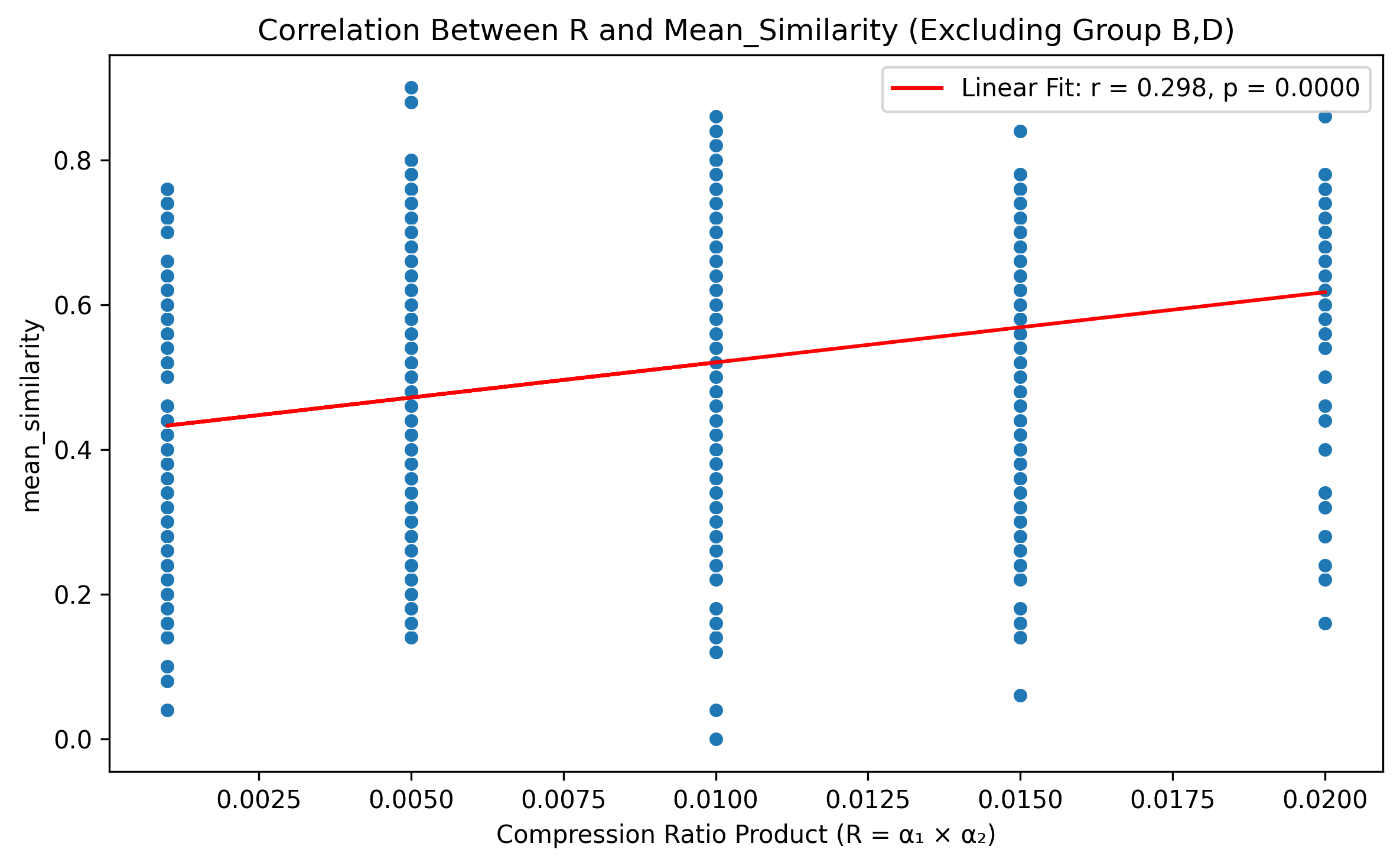}
    \caption{ Correlation between the compression ratio product ($R = \alpha_1 \times \alpha_2$) and the average similarity score, computed as the mean of semantic similarity, plot similarity, character similarity, background similarity, and style similarity, excluding the baseline group B,D. A weak but statistically significant positive correlation is observed ($r = 0.129$, $p < 0.001$).}
    \label{fig:r-vs-mean-similarity}
\end{figure}

\clearpage
\section{Appendix C： Prompts Templates： LongWriter baseline}
\label{se:Appendixbaseline}
\begin{figure}[h]
\centering
\begin{tabular}{|p{\textwidth}|}
\hline
\textbf{Chinese Prompt Template (LongWriter Reconstruction):} \\
根据以下【小说大纲】写第\{chap\_num\}章，要求： \\
1. 字数在5000字。 \\
2. 文笔流畅，语言生动。 \\
\\
【小说大纲】: \{outline\_text\} \\
\\
【第\{chap\_num\}章】正文开始： \\
\hline
\textbf{English Translation:} \\
Based on the following [NOVEL OUTLINE], write Chapter \{chap\_num\}. \\
Requirements: \\
1. Word count: approximately 5,000 words. \\
2. Maintain fluent writing style and vivid language. \\
\\
【NOVEL OUTLINE】: \{outline\_text\} \\
\\
【Chapter \{chap\_num\}】 begins: \\
\hline
\end{tabular}
\caption{Prompt templates used for LongWriter baseline reconstruction. English Translation is used for novels written in English. Chinese Translation is used for novels written in Chinese.}
\label{fig:longwriter_prompts}
\end{figure}

\clearpage

\section{Appendix D： Prompts Templates： Hierarchical Prompts： K=2 Stage 1 Compression}
\label{se:AppendixHierarchicalPrompts}
\begin{figure}[h]
\centering
\begin{tabular}{|p{\textwidth}|}
\hline
\textbf{Chinese Prompt Template (Chapter Analysis):} \\
你是一位专业的文学编辑。请仔细阅读我提供的【章节全文】。 \\
你的任务是提取结构化信息，并 **必须** 调用 `extract\_chapter\_details` 函数来返回结果。 \\
请严格按照函数参数的描述（特别是关于"情节摘要导语"的详细程度要求）来填充信息。 \\
**绝对不要** 输出任何 JSON 格式之外的文本、解释、代码块标记（如 ```json ... ```）或 Markdown。 \\
直接调用函数并填充其参数。 \\
\\
【章节全文】：\{chapter\_text\} \\
\hline
\textbf{English Translation:} \\
You are a professional literary editor. Please carefully read the [CHAPTER TEXT] I provide. \\
Your task is to extract structured information and **must** call the `extract\_chapter\_details` function to return results. \\
Please strictly follow the function parameter descriptions (especially the detailed requirements for "plot summary introduction") to fill in the information. \\
**Never** output any text, explanations, code block markers (like ```json ... ```) or Markdown outside of JSON format. \\
Directly call the function and fill in its parameters. \\
\\
【CHAPTER TEXT】: \{chapter\_text\} \\
\hline
\textbf{JSON Schema Structure:} \\
\texttt{\{} \\
\texttt{~~"情节摘要导语": "string", // Plot summary introduction} \\
\texttt{~~"出现人物": ["string"], // Characters appearing} \\
\texttt{~~"出现道具": ["string"], // Props appearing} \\
\texttt{~~"出现场景": ["string"], // Scenes appearing} \\
\texttt{~~"伏笔\_设下": ["string"], // Foreshadowing set} \\
\texttt{~~"伏笔\_回收": ["string"] // Foreshadowing resolved} \\
\texttt{\}} \\
\hline
\end{tabular}
\caption{Prompt templates used for our hierarchical method's K=2 stage 1 compression to section outline. The JSON schema ensures consistent extraction of narrative elements across all experimental conditions.}
\label{fig:hierarchical_prompts}
\end{figure}

\clearpage
\section{Appendix E： Prompts Templates： Hierarchical Prompts： K=2 Stage 2 Compression}
\begin{figure}[h]
\centering
\begin{tabular}{|p{\textwidth}|}
\hline
\textbf{Chinese Prompt Template (Compression K=2 Stage2):} \\
你是一位资深中文小说编辑，请将下面的【整书细纲】精炼为总字数约10000字的大纲。 \\
【输出要求】 \\
- 只用简体中文，以第几章节：大纲情节的格式输出，不要章节这两个字，保留章节号； \\
- 保持情节完整，突出主要人物、冲突、转折、结局； \\
- 不得输出任何额外解释或注释。 \\
- 每个段落要精简，必须涵盖从第一章到最后一章的全部细纲内容。 \\
约10000字，及每章约40字。200章总结完即停止输出。 \\
\\
【整书细纲】：\{detailed\_outline\} \\
\hline
\textbf{English Translation:} \\
You are a senior Chinese novel editor. Please refine the following [DETAILED BOOK OUTLINE] into an outline of approximately 10,000 words. \\
【Output Requirements】 \\
- Use only Simplified Chinese, output in the format "Chapter X: plot outline", omit the word "章节", keep chapter numbers; \\
- Maintain plot integrity, highlight main characters, conflicts, turning points, and endings; \\
- Do not output any additional explanations or annotations. \\
- Each paragraph should be concise and must cover all detailed outline content from the first chapter to the last chapter. \\
Approximately 10,000 words, about 40 words per chapter. Stop output after summarizing 200 chapters. \\
\\
【DETAILED BOOK OUTLINE】: \{detailed\_outline\} \\
\hline
\end{tabular}
\caption{Prompt templates used for hierarchical compression stage (K=2, Stage2). This template compresses detailed chapter outlines into a concise 10,000-word summary while preserving narrative structure. 10,000 words is a variable.}
\label{fig:compression_prompts}
\end{figure}

\clearpage
\section{Appendix F: Prompts Templates: Hierarchical Prompts: K=2 Stage 2 Expansion}
\begin{figure}[h]
\centering
\begin{tabular}{|p{\textwidth}|}
\hline
\textbf{Chinese Prompt Template (Expansion K=2 Stage2):} \\
请根据大纲的对应章节写一段 200-300 字的情节摘要导语，字数必须在这个区间，不得超出也不得少于。结尾不要加字数统计 \\
你是一位专业中文小说策划。下面给出【整书章节大纲】。 \\
请扩写并输出 **第 \{n\} 章** 的结构化细纲。 \\
必须调用函数 `extract\_chapter\_details` 按参数要求返回结果， \\
绝不能输出 JSON 之外的任何文字或 Markdown。 \\
\\
【整书章节大纲】：\{outline\} \\
\hline
\textbf{English Translation:} \\
Please write a 200-300 word plot summary introduction for the corresponding chapter in the outline. The word count must be within this range, neither exceeding nor falling short. Do not add word count statistics at the end. \\
You are a professional Chinese novel planner. The [COMPLETE BOOK CHAPTER OUTLINE] is given below. \\
Please expand and output the structured detailed outline for **Chapter \{n\}**. \\
Must call the function `extract\_chapter\_details` to return results according to parameter requirements, \\
Never output any text or Markdown outside of JSON. \\
\\
【COMPLETE BOOK CHAPTER OUTLINE】: \{outline\} \\
\hline
\textbf{JSON Schema Structure:} \\
\texttt{\{} \\
\texttt{~~"情节摘要导语": "string", // Plot summary introduction (400-500 words)} \\
\texttt{~~"出现人物": ["string"], // Characters appearing} \\
\texttt{~~"出现道具": ["string"], // Props appearing} \\
\texttt{~~"出现场景": ["string"], // Scenes appearing} \\
\texttt{~~"伏笔\_设下": ["string"], // Foreshadowing set} \\
\texttt{~~"伏笔\_回收": ["string"] // Foreshadowing resolved} \\
\texttt{\}} \\
\hline
\end{tabular}
\caption{Prompt templates used for hierarchical expansion stage (K=2, Stage2). This template expands compressed outlines into structured 50,000-word detailed chapter outlines with specific narrative elements. The 200-300 word requirement ensures around 50,000 words in total.}
\label{fig:expansion_prompts}
\end{figure}

\clearpage
\section{Appendix G: Prompts Templates: K=1 Direct Compression Method}

\begin{figure}[h!]
\centering
\begin{tabular}{|p{\textwidth}|}
\hline
\textbf{Chinese Prompt Template (Direct Compression):} \\
你是一位资深中文小说编辑，现在需要为整本书撰写一份【整书大纲】。 \\
\\
【输出规范】 \\
1. 全文仅使用简体中文； \\
2. 字数 ≤1000 汉字； \\
3. 完整概括出主线剧情故事，尤其是主要人物、核心冲突、关键转折与结局； \\
4. 不要出现章节标题、序号、列表符号，直接以自然段叙述； \\
5. 开头不得使用"以下是"或类似提示语，应直接进入正文。 \\
6. 概括全文，注意，是概括100万字小说的从开头到结尾的故事。 \\
7. 概括全文的同时保留尽量多的细节，尽量多的人物，尽量多的重要情节 \\
8. 告诉我最后一章节的标题，这个部分不算在1000字的限制内，作为你阅读了整本书的测试 \\
\\
请严格遵守以上规则，一次性输出完成后的整书概要。 \\
\\
【完整小说文本】：\{full\_novel\_text\} \\
\hline
\textbf{English Translation:} \\
You are a senior Chinese novel editor, and now you need to write a [COMPLETE BOOK OUTLINE] for the entire book. \\
\\
【Output Specifications】 \\
1. Use only Simplified Chinese throughout; \\
2. Word count ≤1000 Chinese characters; \\
3. Completely summarize the main storyline, especially main characters, core conflicts, key turning points and endings; \\
4. Do not include chapter titles, numbers, or list symbols, narrate directly in natural paragraphs; \\
5. Do not start with "The following is" or similar prompts, should directly enter the main text. \\
6. Summarize the full text, note that this is to summarize a 1 million word novel from beginning to end. \\
7. While summarizing the full text, retain as many details, characters, and important plots as possible \\
8. Tell me the title of the last chapter, this part does not count towards the 1000-word limit, as a test of your reading of the entire book \\
\\
Please strictly follow the above rules and output the completed book summary in one go. \\
\\
【COMPLETE NOVEL TEXT】: \{full\_novel\_text\} \\
\hline
\end{tabular}
\caption{Prompt templates used for direct compression method. This approach directly compresses the complete novel (1 million words) into a concise 1000-word outline while preserving essential narrative elements, characters, and plot details. The last chapter title requirement serves as a verification mechanism.}
\label{fig:direct_compression_prompts}
\end{figure}

\clearpage
\section{Appendix H: Prompts Templates: K=2 Mixed Hierarchical Direct Expansion and K=1 Expansion}
\begin{figure}[h]
\centering
\begin{tabular}{|p{\textwidth}|}
\hline
\textbf{Chinese Prompt Template (Mixed Hierarchical Direct Expansion):} \\
你是一位擅长情节创作的中文作家，现在需要根据【整书大纲】扩写第 \{chap\_num\} 章。 \\
\\
【整书大纲】 \\
\{outline\_text\} \\
\\
【写作要求】 \\
1. 语言生动连贯； \\
2. 字数绝对不要少于 5000 字； \\
3. 聚焦本章情节； \\
4. 只输出正文，无标题。 \\
\hline
\textbf{English Translation:} \\
You are a Chinese writer skilled in plot creation. Now you need to expand Chapter \{chap\_num\} based on the [COMPLETE BOOK OUTLINE]. \\
\\
【COMPLETE BOOK OUTLINE】 \\
\{outline\_text\} \\
\\
【Writing Requirements】 \\
1. Vivid and coherent language; \\
2. Word count must not be less than 5000 words; \\
3. Focus on this chapter's plot; \\
4. Output only the main text, no title. \\
\hline
\end{tabular}
\caption{Prompt templates used for mixed hierarchical direct expansion method. This approach directly expands from compressed outline to full chapter content (5000+ words) without intermediate structured analysis, providing a streamlined generation process while maintaining narrative quality.}
\label{fig:mixed_hierarchical_prompts}
\end{figure}

\clearpage
\section{Appendix I: Prompts Templates: LLM-based Evaluation}

\begin{figure}[h!]
\centering
\begin{tabular}{|p{\textwidth}|}
\hline
\textbf{Chinese Prompt Template (LLM Evaluation):} \\
你是一位专业中文小说编辑，请你阅读【文本A】与【文本B】，完成以下任务： \\
1. 分别提取文本A与文本B中的： \\
~~~- 出现道具列表（如：剑、玉、令牌等） \\
~~~- 出现人物名称列表 \\
~~~- 出现场景/环境名称列表 \\
~~~* 提取时请尽量精确，去除通用词语 (例如: '人', '地方')，只保留具体名称。 \\
~~~* 如果某一项在文本中没有出现，请返回空列表 `[]`。 \\
2. 分别统计每类元素的数量（去重后），并输出每类的元素列表与数量。 \\
3. 接着对比两段文本内容，按照以下 5 个维度进行 0-1 评分（1 表示非常相似，0 表示完全不同）： \\
~~~- semantic\_similarity~~~整体语义/主题 \\
~~~- plot\_similarity~~~~~~~情节、事件发展 \\
~~~- character\_similarity~~人物名称、数量与设定（综合考虑） \\
~~~- background\_similarity~场景与世界设定 \\
~~~- style\_similarity~~~~~~语言风格与表达方式 \\
~~~* 评分请基于文本内容，给出客观评估。 \\
**请严格输出以下 JSON 格式，不要包含 markdown ```json ... ``` 标记，直接输出 JSON 对象：** \\
【文本A】\{text\_a\} \\
【文本B】\{text\_b\} \\
\hline
\textbf{English Translation:} \\
You are a professional Chinese novel editor. Please read [TEXT A] and [TEXT B] and complete the following tasks: \\
1. Extract from Text A and Text B respectively: \\
~~~- List of props appearing (e.g., sword, jade, token, etc.) \\
~~~- List of character names appearing \\
~~~- List of scene/environment names appearing \\
~~~* Please extract as precisely as possible, remove generic words (e.g., 'person', 'place'), keep only specific names. \\
~~~* If any category does not appear in the text, please return empty list `[]`. \\
2. Count the number of each type of element (after deduplication) and output the element list and count for each category. \\
3. Then compare the two text contents and score on the following 5 dimensions from 0-1 (1 means very similar, 0 means completely different): \\
~~~- semantic\_similarity~~~Overall semantics/theme \\
~~~- plot\_similarity~~~~~~~Plot and event development \\
~~~- character\_similarity~~Character names, quantity and settings (comprehensive consideration) \\
~~~- background\_similarity~Scenes and world settings \\
~~~- style\_similarity~~~~~~Language style and expression \\
~~~* Please give objective evaluation based on text content. \\
**Please strictly output the following JSON format, do not include markdown ```json ... ``` markers, output JSON object directly:**
\\
【TEXT A】 \{text\_a\} \\
【TEXT B】 \{text\_b\} \\
\hline

\hline
\end{tabular}
\caption{Prompt templates used for LLM-based evaluation across five similarity
dimensions: semantic, plot, character, background, and
style.}
\label{fig:llm_evaluation_prompts}
\end{figure}

\clearpage
\section{Appendix J: Sampling Design}

\label{sec:design}

We implement a two-stage sampling design :
%


We implement a two-stage sampling design:

\textbf{Stage 1 (between-novel):}
We stratify the sampling frame into four major genres—Urban (U), Romance (R), Fantasy (F), and Historical (H)—and select \(n_h\) novels from each stratum \(h\) using probability-proportional-to-size (PPS) sampling, where the size variable is the total word count \(L_i \approx 1\) million. The total sample size is set to \(n = \mathbf{40}\), with allocation determined by Neyman’s optimal allocation scheme \citep{Olayiwola2013Neyman}:
\[
    n_h = n\,\frac{N_h S_h}{\sum_{g} N_g S_g},
\]
where \(N_h\) and \(S_h\) denote the number of novels and estimated standard deviation of the distortion metric within stratum \(h\), respectively.

\textbf{Stage 2 (within-novel):}
For each selected novel \(i\), we treat entire chapters as secondary sampling units. We sample \(m_i = \mathbf{8}\) chapters using simple random sampling without replacement (SRSWOR). If a novel contains fewer than 8 chapters, all chapters are included.

In this study, we do not incorporate sampling weights; all selected units are treated equally.

\end{CJK}
\end{document}